\relax
\documentclass[letterpaper]{article} % DO NOT CHANGE THIS
\usepackage{aaai20}  % DO NOT CHANGE THIS
\usepackage{times}  % DO NOT CHANGE THIS
\usepackage{helvet} % DO NOT CHANGE THIS
\usepackage{courier}  % DO NOT CHANGE THIS
\usepackage[hyphens]{url}  % DO NOT CHANGE THIS
\usepackage{graphicx} % DO NOT CHANGE THIS
\usepackage{graphicx}  % DO NOT CHANGE THIS
\usepackage{amsmath} % assumes amsmath package installed
\usepackage{amssymb}  % assumes amsmath package installed
\usepackage{algpseudocode,algorithm,algorithmicx}
\usepackage{multirow}
\usepackage{bm}
\usepackage{booktabs}
\usepackage{array}
\usepackage{amsfonts}
\usepackage{adjustbox}
\usepackage{gensymb} % math degree symbol
\usepackage{subcaption}
\usepackage{threeparttable}
\DeclareMathSizes{10}{8.5}{6.5}{6.5}

\makeatletter
\newcommand{\printfnsymbol}[1]{%
  \textsuperscript{\@fnsymbol{#1}}%
}
\makeatother

\makeatletter
\def\@fnsymbol#1{\ensuremath{\ifcase#1\or \dagger\or \ddagger\or
   \mathsection\or \mathparagraph\or \|\or **\or \dagger\dagger
   \or \ddagger\ddagger \else\@ctrerr\fi}}
\makeatother

\title{RoboCoDraw: Robotic Avatar Drawing with GAN-based Style Transfer and Time-efficient Path Optimization}

\author{Tianying Wang\textsuperscript{\rm 1,}\thanks{\;The first three authors contributed equally.},
Wei Qi Toh\textsuperscript{\rm 1,}\printfnsymbol{1}, 
Hao Zhang\textsuperscript{\rm 1,}\printfnsymbol{1} \\ 
\Large \textbf{Xiuchao Sui\textsuperscript{\rm 1}, 
Shaohua Li\textsuperscript{\rm 1,2}, 
Yong Liu\textsuperscript{\rm 1,2}, 
Wei Jing\textsuperscript{\rm 1,3,}}\thanks{\;Corresponding author.}\\ 
\textsuperscript{\rm 1}Artificial Intelligence Initiative, A*STAR\\
\textsuperscript{\rm 2}Institute of High Performance Computing, A*STAR\\
\textsuperscript{\rm 3}Institute of Information Research, A*STAR\\
1 Fusionopolis Way, Connexis North Tower, 138632, Singapore\\
21wjing@gmail.com % email address must be in roman text type, not monospace or sans serif
}
\begin{document}

\maketitle

\begin{abstract}

Robotic drawing has become increasingly popular as an entertainment and interactive tool. In this paper we present RoboCoDraw, a real-time collaborative robot-based drawing system that draws stylized human face sketches interactively in front of human users, by using the Generative Adversarial Network (GAN)-based style transfer and a Random-Key Genetic Algorithm (RKGA)-based path optimization. The proposed RoboCoDraw system takes a real human face image as input, converts it to a stylized avatar, then draws it with a robotic arm. A core component in this system is the AvatarGAN proposed by us, which generates a cartoon avatar face image from a real human face. AvatarGAN is trained with unpaired face and avatar images only and can generate avatar images of much better likeness with human face images in comparison with the vanilla CycleGAN. After the avatar image is generated, it is fed to a line extraction algorithm and converted to sketches. An RKGA-based path optimization algorithm is applied to find a time-efficient robotic drawing path to be executed by the robotic arm. We demonstrate the capability of RoboCoDraw on various face images using a lightweight, safe collaborative robot UR5.
\end{abstract}

\begin{figure}[!htbp]
      \centering
      \includegraphics[width=0.45\textwidth]{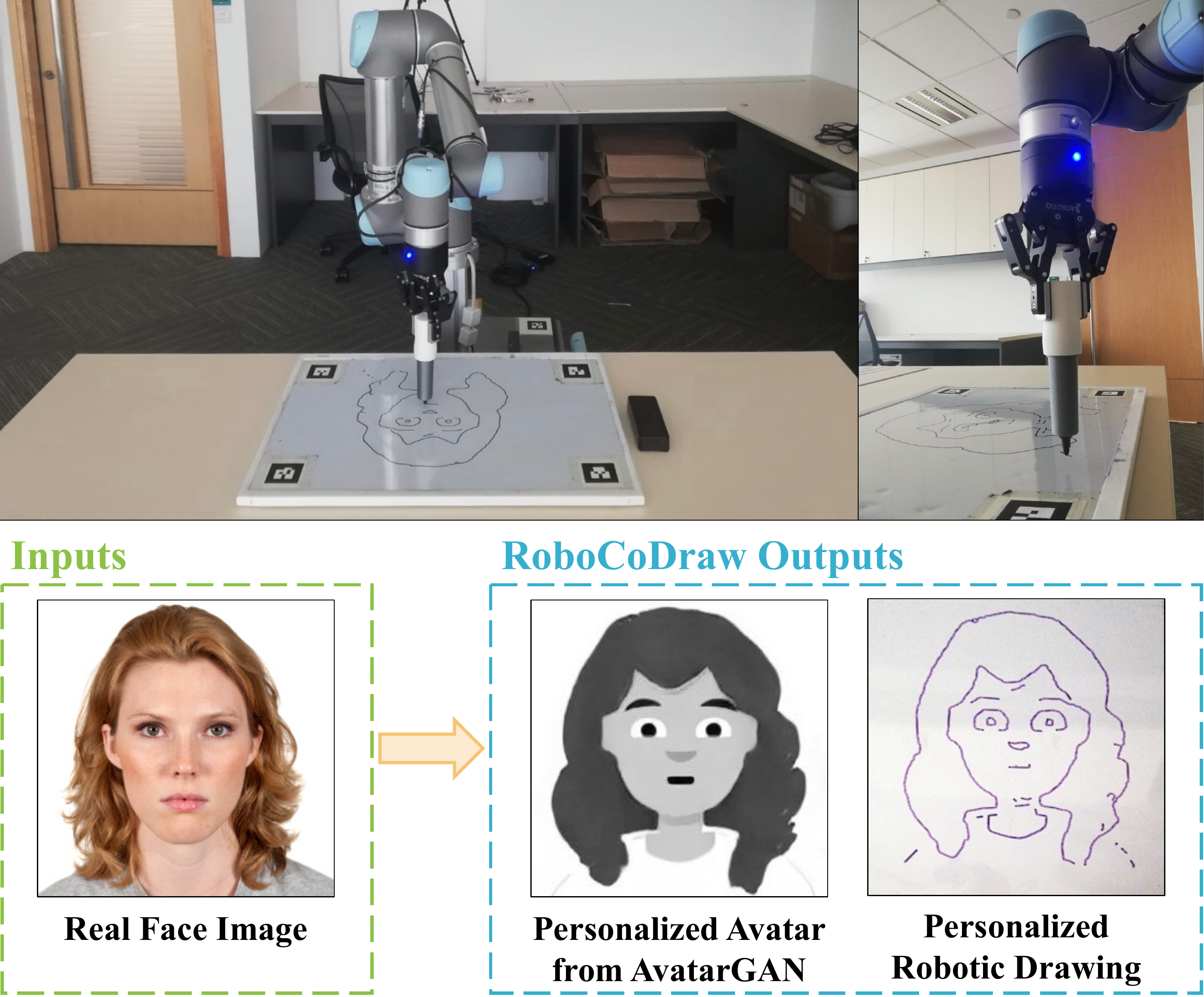}
        \caption{Top: the robot performing the drawing; Bottom: the generated personalized avatar and the drawing result.}
      \label{fig:abstract_robocodraw} 
\end{figure} 

\section{Introduction}

Robotic drawing is an increasingly popular human-robot interaction task that is fascinating and entertaining to the public. In a typical robotic drawing system, a robotic arm \cite{song2018artistic} or a humanoid robot \cite{calinon2005humanoid} is usually used in an interactive environment to draw pictures in front of human users.
Because of its interactivity and entertainment, robotic drawing systems have found applications in a wide range of scenarios, such as early childhood education \cite{hood2015children}, psychological therapy \cite{cooney2018design}, and social entertainment \cite{jean2012artist}. Among these different systems, drawing human faces is one of the most engaging and entertaining tasks. During the past years, robotic face drawing has been extensively studied \cite{calinon2005humanoid}, with a focus on generating realistic portraits.

Although somewhat impressive, realistic drawings have a limited level of amusement for human users. In order to increase entertainment and engagement, in this paper, we propose a real-time robotic drawing system that converts input face images to cartoon avatars as a form of robotic art creation. Our system can capture and preserve facial features of the input face images and reproduce them in generated avatars so that the avatars have good likeness with the input faces. Such personalized art creation enriches the interactive experience between the user and the robot, while the optimized execution of the robotic drawing complements and improves the overall experience. Additionally, the whole system is trained without manual annotations.

\begin{figure*}[t]
  \centering
  \includegraphics[width=0.9\textwidth]{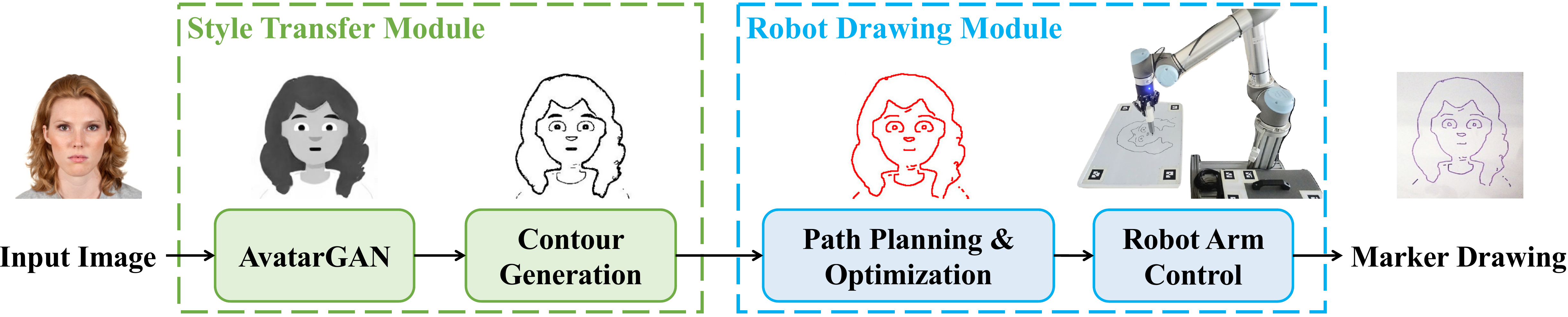}
  \caption{Pipeline of the RoboCoDraw System.}
  \label{fig:codrawsystem}
\end{figure*}

In this paper, we propose \textit{RoboCoDraw}, a collaborative robot-based, real-time drawing system. The core component of RoboCoDraw is \emph{AvatarGAN}, a Generative Adversarial Network (GAN)-based image style transfer method for transferring real human faces to personalized cartoon avatars for robotic drawing. After generating the avatar with AvatarGAN, our system utilizes a line extraction algorithm to convert the avatar to a sketch. Subsequently, a Random-Key Genetic Algorithm (RKGA)~\cite{RKGA1994} based optimization algorithm is adopted to find a time-efficient robotic path for completing the avatar drawing.

Our main contributions are:
\begin{itemize}
    \item a two-stream CycleGAN model named AvatarGAN, that maps human faces to cartoon avatars, while preserving facial features (such as haircut, face shape, eye shape, face color); 
    \item a modularized, interactive robotic drawing system that performs faithful style translation and time-efficient face drawing; and
    \item a path optimization formulation for the robotic drawing problem and an RKGA-based optimization algorithm with two-level local improvement heuristics, so as to find time-efficient robot paths.
\end{itemize}

\section{Relevant Work}

The proposed \textit{RoboCoDraw} system consists of two modules: a style transfer module and a robotic drawing path planning module. The related work to these two modules is discussed separately.

\subsection{Style Transfer}

A lot of research work has been carried out on style transfer with different approaches, such as neural style loss-based methods \cite{li2017laplacian} and Generative Adversarial Network \cite{goodfellow2014generative} (GAN)-based methods. The GAN-based methods are recent approaches that achieve high-quality style transfer using relatively small datasets. For example, CycleGAN~\cite{cyclegan2017} performs image style translation using unpaired image datasets, which yields good texture mapping between objects with similar geometries, such as horse-to-zebra and orange-to-apple. However, it is not good at capturing facial features that are to be used to generate personalized avatars with a good likeness of input faces, as shown in our experiments.

CariGANs~\cite{cao2018cari} are another approach for photo-to-caricature translation, which consists of two networks, CariGeoGAN and CariStyGAN. CariGeoGAN performs geometrical transformation while CariStyGAN focuses on style appearance translation. Although they result in vivid caricatures, the caricatures come with a lot of details and are not well suited to our application scenario of quick drawing with robotic arms.

Moreover, it requires manually labeled facial landmarks, both on real face images and on caricatures, to train CariGeoGAN.

\subsection{Path Planning for Robotic Drawing}

The path planning module aims to reduce the time required for physical execution of the drawing. There are several approaches for planning robot drawing paths from a reference image, one of which is path generation based on replication of pixel shade and intensities in the reference image \cite{calinon2005humanoid,paul2013}.  The Travelling Salesman Problem (TSP) line art combines such an approach with path optimization for efficient continuous line drawings \cite{bridges2005:301}. However, the final drawings obtained from replicating pixel shades lack the aesthetics and attractiveness of line art. Another work involving robotic drawing path planning attempted to reduce the number of output points and candidate trajectories \cite{lin2009human}, but the approach was ad-hoc and lacked quantitative evaluation. 

Additional research has been done on robotic path optimization and robotic task sequencing, as summarized in~\cite{Alatartsev2015RoboticTS}, which could be applied to the robot drawing problem. Task sequencing, i.e., choosing an efficient sequence of tasks and the way the robot is moving between them, can be formulated as a TSP or its variants. It is often done manually or with offline path planning using a greedy search algorithm ~\cite{Alatartsev2015RoboticTS} in industrial robotic applications. Other tour-searching algorithms that can be used for solving sequencing problems include Genetic Algorithms~\cite{gataskseq2005}, Ant Colony optimization~\cite{YANG20081417}, and variety of local search heuristics~\cite{Lin1973}.

For our application, the robotic drawing problem is formulated as a Generalized Travelling Salesman Problem (GTSP), a variant of TSP. The proposed formulation models the path optimization problem more accurately and thus yields better results.

\section{The RoboCoDraw System}

In this section, we introduce the proposed \textit{RoboCoDraw} system in detail. We first present the style transfer module, whose core component is \textit{AvatarGAN}, a novel GAN-based style transfer model for the real-face image to avatar translation; then we describe the robotic drawing module that performs planning and optimization of the drawing paths. Fig. \ref{fig:codrawsystem} shows the pipeline of the \textit{RoboCoDraw} system.

\subsection{Style Transfer Module}

The style transfer module consists of two components. One is \textit{AvatarGAN}, which performs photo-to-avatar translation; the other is \textit{Contour Generation} for extracting coherent contours from generated avatars. The two components work in sequence to produce stylized avatar sketches based on real face images.

\subsubsection{AvatarGAN}
\begin{figure}[!htpb]
      \centering
      \includegraphics[ width=0.44\textwidth]{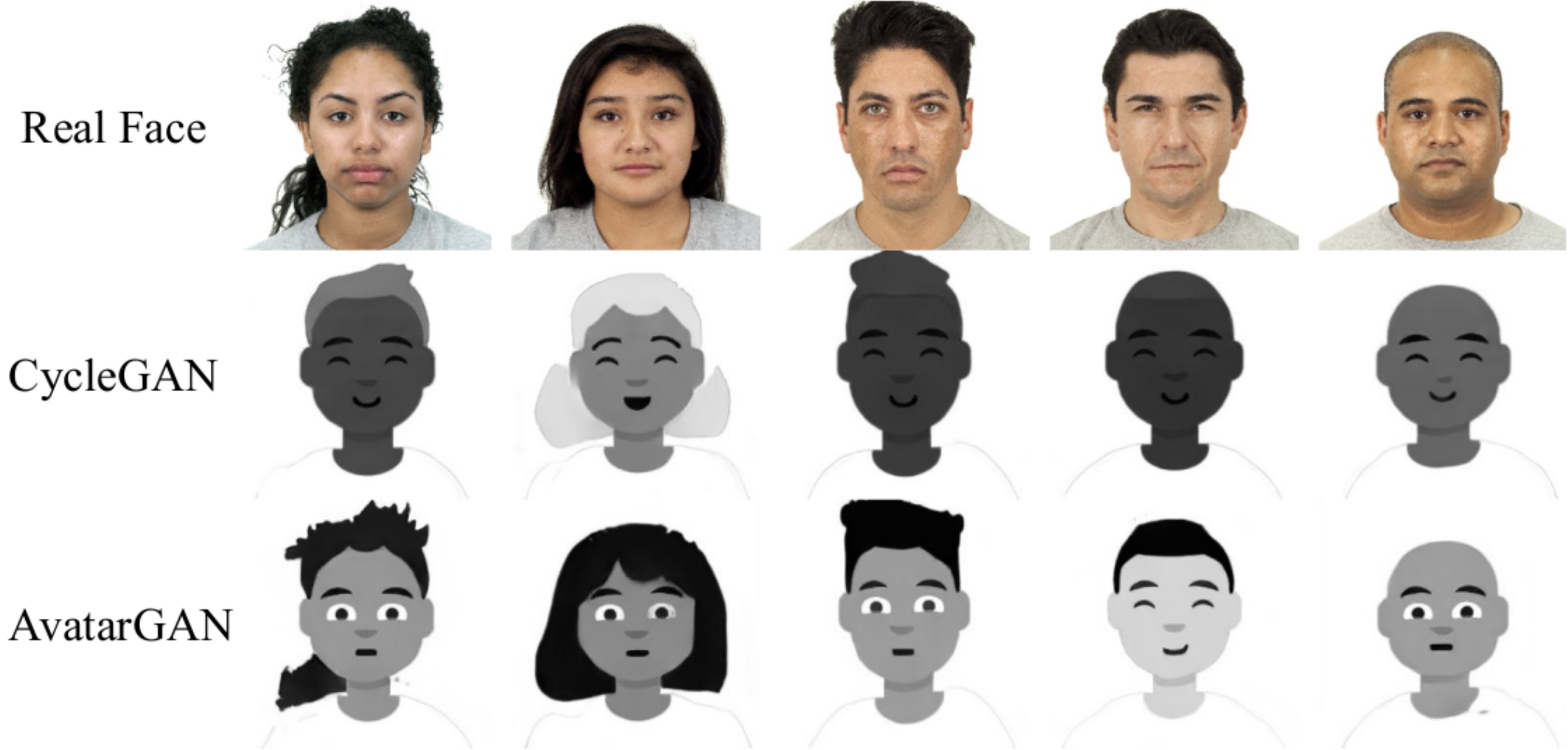}
      \caption{Examples of real face to cartoon-style avatar translation using CycleGAN and AvatarGAN.}
      \label{fig:result_example}
\end{figure}

The objective of AvatarGAN is to learn a mapping between the domain of real faces  ($\textit{X}$) and the domain of cartoon-style avatars ($\textit{Y}$) while preserving the consistency in facial features such as haircut, face color, face shape, and eye shapes.

In order to make an approach broadly applicable to different avatar datasets, it is desired to be fully unsupervised. CycleGAN is an effective approach to learn the domain mapping in a fully unsupervised manner. However, the vanilla CycleGAN fails to preserve the facial features of real faces in the generated avatars (ref. Fig.~\ref{fig:result_example}), since it aims to learn the mapping between whole images in the two domains, without focusing on any specific areas within the images. To address this problem, we propose a two-stream CycleGAN model named AvatarGAN, in which an additional stream focuses on translating facial areas only. The two-stream architecture forces AvatarGAN to preserve important facial features when translating the whole image.

\begin{figure}[!htpb]
      \centering
      \includegraphics[width=0.45\textwidth]{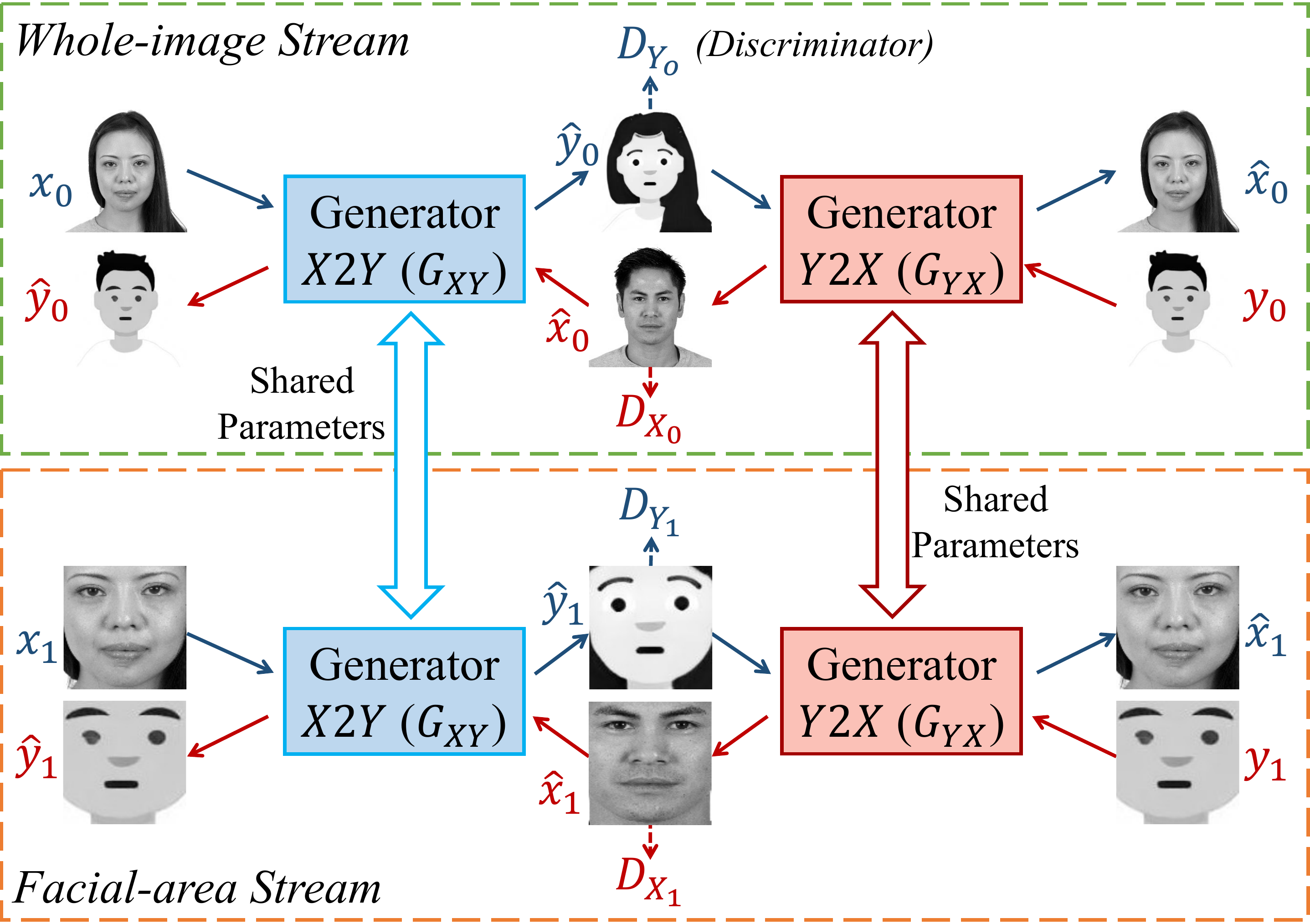}
      \caption{The structure of AvatarGAN, which consists of two translation streams. The whole-image stream performs ordinary style translation between real faces and avatars, while the facial-area stream focuses on learning facial feature mappings between the two domains.}
      \label{fig:avatar_gan}
\end{figure}

The structure of AvatarGAN is shown in Fig.~\ref{fig:avatar_gan}. The training real face images are denoted as $\{x_{0,i}\}_{i=1}^{N} \subset \textit{X}_0$, the training avatar images as $\{y_{0,j}\}_{j=1}^{M} \subset \textit{Y}_0$, the cropped real facial area images as $\{x_{1,i}\}_{i=1}^{N} \subset \textit{X}_1$, and the cropped avatar facial area images as $\{y_{1,j}\}_{j=1}^{M} \subset \textit{Y}_1$.

Let the whole-image data distribution be $x\sim p_0(x)$ and $y\sim p_0(y)$, and the facial-area data distribution be $x\sim p_1(x)$ and $y\sim p_1(y)$.
In the whole-image translation stream, the generator $G_{X_0 Y_0}$ performs the $\textit{X}_0 \to \textit{Y}_0$ translation (from real face domain $\textit{X}_0$ to avatar domain $\textit{Y}_0$), and $G_{Y_0 X_0}$ performs $\textit{Y}_0 \to \textit{X}_0$ translation. In the facial-area translation stream, the two generators $G_{X_1 Y_1}$ and $G_{Y_1 X_1}$ are similarly defined.  

To make the translations in the two streams produce images in consistent styles, we make the weights shared by the generators in the two streams, i.e., $G_{X_0 Y_0} = G_{X_1 Y_1}$, and $G_{Y_0 X_0} = G_{Y_1 X_1}$. For convenience, we drop the subscript and denote them as $G_{XY}$ and $G_{YX}$.
In addition, we introduce four adversarial discriminators $D_{\textit{X}_0}$, $D_{\textit{X}_1}$, $D_{\textit{Y}_0}$, and $D_{\textit{Y}_1}$. They aim to predict the probabilities that an image is from domain $\textit{X}_0$, $\textit{X}_1$, $\textit{Y}_0$ and $\textit{Y}_1$, respectively.

The objectives of AvatarGAN are: 1) minimize the adversarial loss ($ \mathcal{L}_{adv}$) on both streams to align the distributions of generated images with the target domains; and 2) minimize the cycle consistency loss ($\mathcal{L}_{cycle}$) on both streams to enforces the learned mappings $G_{\textit{XY}}$ and $G_{\textit{YX}}$ to form inverse mappings with each other.

For adversarial loss, we only express the definition of the whole-image stream, and that on the facial-area stream is similar. Given the generator $G_{\textit{XY}}$ and its corresponding discriminator $D_{\textit{Y}_0}$, the adversarial loss is defined as:
\begin{equation}
\begin{split}
    \mathcal{L}_{adv}(G_{\textit{XY}},D_{\textit{Y}_0},\textit{X}_0,\textit{Y}_0)
    &=\mathbb{E}_{y\sim p_0(y)}[\log D_{\textit{Y}_0}(y)] \\
    &+\mathbb{E}_{x\sim p_0(x)}[\log(1- D_{\textit{Y}_0}(G_{\textit{XY}}(x))].
\end{split}
\end{equation}
Similarly, we have $ \mathcal{L}_{adv}(G_{\textit{YX}},D_{\textit{X}_0},\textit{X}_0,\textit{Y}_0)$ for the generator $G_{\textit{YX}}$ and its discriminator $D_{\textit{X}_0}$. The adversarial losses on the facial-area stream  $\mathcal{L}_{adv}(G_{\textit{XY}},D_{\textit{Y}_1},\textit{X}_1,\textit{Y}_1)$ and $\mathcal{L}_{adv}(G_{\textit{YX}},D_{\textit{X}_1},\textit{X}_1,\textit{Y}_1)$ for mappings $G_{\textit{XY}}$ and $G_{\textit{YX}}$ are defined in the same manner, respectively. The four adversarial losses are summed up as the overall adversarial loss ($\mathcal{L}_{adv}$).

\begin{figure}[thpb]
      \centering
      \includegraphics[width=0.45\textwidth]{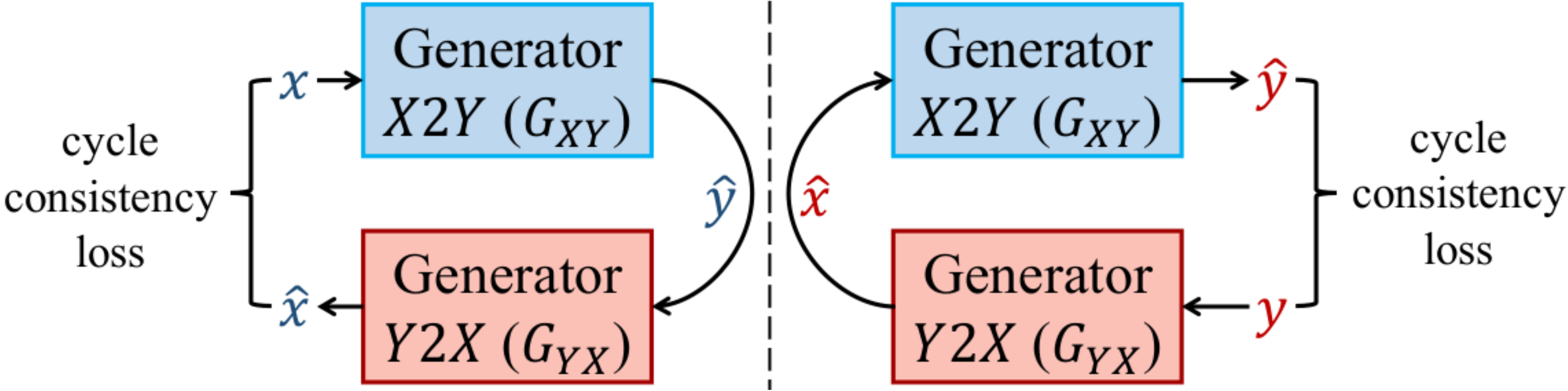}
      \caption{The cycle consistency loss, which enforces the learned mappings $G_{\textit{XY}}$ and $G_{\textit{YX}}$ to form inverse mappings.}
      \label{fig:cycle_consistency}
\end{figure}

The cycle consistency loss (Fig.~\ref{fig:cycle_consistency}) is defined to constrain the learned mapping functions to be cycle-consistent. In the whole-image translation stream, for each image $x$ from domain $\textit{X}_0$, an image translation cycle is expected to convert $x$ back to the original image, i.e. $x \to G_{\textit{XY}}(x) \to G_{\textit{YX}}(G_{\textit{XY}}(x)) \approx x$; while for each image $y$ from domain $\textit{Y}_0$, a translation cycle is also expected to satisfies $y \to G_{\textit{YX}}(y) \to G_{\textit{XY}}(G_{\textit{YX}}(y)) \approx y$. Thus, the cycle consistency loss in the whole-image stream is defined as:
\begin{equation}
\begin{split}
    \mathcal{L}_{cycle}(G_{\textit{XY}},G_{\textit{YX}}, X_0, Y_0)
    &=\mathbb{E}_{x\sim p_0(x)}\big[\|G_{\textit{YX}}(G_{\textit{XY}}(x))-x\|_{1}\big]\\
    &+\mathbb{E}_{y\sim p_0(y)}\big[\|G_{\textit{XY}}(G_{\textit{YX}}(y))-y\|_{1}\big].
\end{split}
\end{equation}

Similarly, the cycle consistency loss on the facial-area translation stream is denoted as $\mathcal{L}_{cycle}(G_{\textit{XY}},G_{\textit{YX}}, X_1, Y_1)$. Then the overall cycle consistency loss is
\begin{equation}
\begin{split}
    \mathcal{L}_{cycle} = \; &\alpha\cdot\mathcal{L}_{cycle}(G_{\textit{XY}},G_{\textit{YX}}, X_0, Y_0) + \\
    &(1-\alpha)\cdot\mathcal{L}_{cycle}(G_{\textit{XY}},G_{\textit{YX}}, X_1, Y_1),
\end{split}
\end{equation}
where $\alpha$ is a hyperparameter to balance the contributions between the whole-image and facial-area streams. 

Finally, the full optimization objective is:
\begin{equation}
    \mathcal{L}=\mathcal{L}_{adv} + \lambda\cdot\mathcal{L}_{cycle},
\end{equation}
where the hyperparameter $\lambda$ assigns the relative importance of the cycle consistency loss within the whole loss.

\subsubsection{Contour Generation}
 In order to make it ready to be drawn by the robotic arm, the generated avatar is processed to generate contours with the Flow-based Difference-of-Gaussian (FDoG) filtering \cite{kang2009flow}. During this process, the Edge Tangent Flow (ETF) is first constructed from the input image; then a DoG filter is applied to the local edge flow in the perpendicular directions. More formally, a one-dimension filter is first applied to each pixel along the line that is perpendicular to ETF, and then accumulate the individual filter responses along the ETF direction.

The FDoG filter only enhances coherent lines and reduces isolated tiny edges,  thereby results in coherent and clean contours. 

\subsection{Robotic Drawing Module}

The robotic drawing module plans an optimized path for the robotic arm to draw the avatars generated from the style transfer module. It first extracts the pixel-coordinates required for the robot drawing path, then optimizes the path by formulating the robotic drawing problem as a Generalized Travelling Salesman Problem (GTSP). 

\subsubsection{Ordered Pixel-Coordinates Extraction}

The image from the style transfer module is cleaned to increase clarity before subsequent processes. The lines of the binary image are further thinned to one-pixel in width for easy line tracing. The thinning process involves skeletonization, pruning, and line-ends extension, and is achieved by the hit-or-miss transform.

\begin{figure}[!ht]
      \centering
      \includegraphics[width=0.45\textwidth]{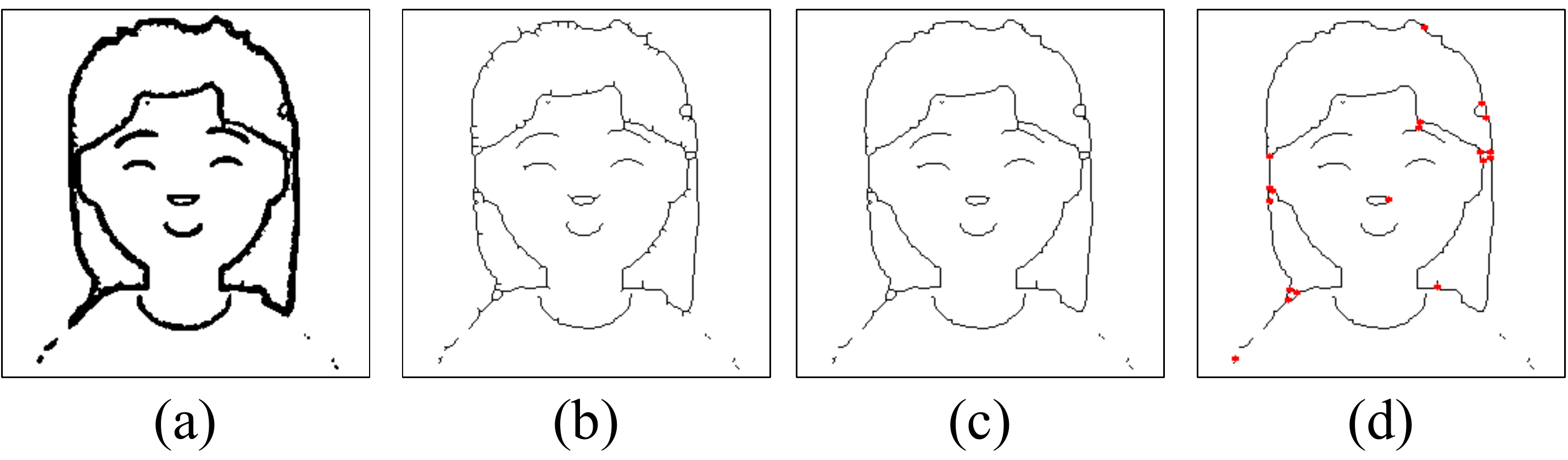}
      \caption{The process of obtaining pixel-coordinates from the reference image. (a) Cleaned image; (b) Image skeleton; (c) Trimmed skeleton; (d) Image split at junctions (in red).}
      \label{fig:junction_detect}
\end{figure}

The thinned image is then split into several line segments from the junction points. The line segments are traced to obtain a sequence of pixel-coordinates to be visited in order when drawing the corresponding line segment. The ordered pixel-coordinates are subsequently stored and passed to the path optimization module to plan a low-cost path. Examples of outputs for each step in the ordered pixel-coordinates extraction process are shown in Figure \ref{fig:junction_detect}.

\subsubsection{Robotic Drawing Path Optimization}

GTSP, a variant of TSP, is applied to formulate the path optimization problem. GTSP partitions the nodes into sets and aims to find the minimum cost tour that visits exactly one node in each set. Thus, both the visiting sequence of the sets as well as the particular nodes to visit need to be determined for solving GTSP \cite{noon1993efficient}. 

In the proposed GTSP, each line segment is represented as a set of two nodes, where each node represents a different drawing direction. The Random-Key Genetic Algorithm (RKGA) is then used to solve the GTSP. In RKGA, the random keys stored in the genes encode the drawing path, while a decoding process recovers the path for fitness evaluation. The encoding/decoding processes of RKGA generally help to map the feasible space to a well-shaped space and convert a constrained optimization problem to an unconstrained one \cite{RKGA1994}.

\begin{figure}[ht]
      \centering
      \includegraphics[width=0.45\textwidth]{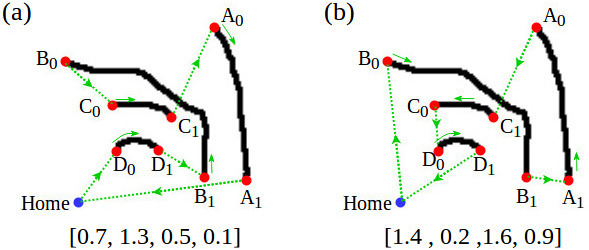}   \caption{Examples of the path encoding of RKGA for robotic drawing applications. Line segments A, B, C and D corresponds to keys 1, 2, 3 and 4, respectively. There are two possible directions for traversing each segment ($0 \to 1$ or $1 \to 0$). The green arrows indicate the drawing path.}
      \label{fig:encode}
\end{figure}

\paragraph{Encoding and decoding with random keys}
The drawing path is encoded as a list of random keys, where the index of each key corresponds to a specific line segment. The keys are real numbers, with the decimal part encoding the visit sequence (line segments corresponding to keys with smaller decimal value are visited earlier in the drawing sequence) and the integer part encoding the direction of drawing the line segment ($1$ represents that the pixel-coordinates traced from the line segment are visited in the forward sequence, while $0$ represents that the pixel-coordinates are visited in reverse). A simplified example of an encoding/decoding path with random keys is shown in Fig. \ref{fig:encode}.

\paragraph{Fitness evaluation}

For robotic drawing, travel time increases with additional distance traversed while the marker is lifted from the whiteboard. Therefore, for a drawing path with $K$ line segments in which $l_1,l_2,...,l_K$ are the lines ordered in visit sequence, the main factors which contribute to the fitness value of the path are: 
\begin{itemize}
    \item the distances $d_{i,i+1}$ from the end of each line $l_i$ to start of the next line $l_{i+1}$ for $i=1,...,K-1$;
    \item the distances to and from the home position for the first and last lines in the tour ($d_{h,1}$ and $d_{K,h}$ respectively);
    \item the number of times the marker is lifted from whiteboard, $n_{\textrm{lift}}$, along with the additional cost $cost_{\textrm{lift}}$ incurred from each time the marker is lifted and placed.
\end{itemize}
 
Assuming constant velocity of the robot end-effector when the marker travels through free space, the increase in travel time is proportional to the increase in distance traveled through free space. Thus, we have
\begin{equation}\label{eq:fitness}
    v_{\textrm{fitness}} = n_{\textrm{lift}}\times cost_{\textrm{lift}} + d_{h,1} + d_{K,h} + \sum_{i=1}^{K-1} d_{i,i+1}.
\end{equation}

The goal of the RKGA is to minimize the fitness/cost value $v_{\textrm{fitness}}$ of the robot drawing the path.

\paragraph{Genetic operators}
We employ three commonly-used genetic operators: reproduction, crossover, and mutation. For the reproduction operation, an elitist strategy is employed to clone the best $r$ individuals directly into the next generation, to ensure the non-degradation of the best solution generated thus far by the algorithm. The remaining $n-r$ individuals for the next generation are produced from a pool of individuals selected via tournament selection from the parent population. Crossover and mutation, which are the second and third genetic operator respectively, are applied to the individuals in the selected pool with probabilities $p_{\textrm{crossover}}$ and $p_{\textrm{mutation}}$. Uniform crossover was used as the crossover strategy, while index shuffling and bit flip were used to mutate the of the decimal part and integer part of the key respectively.

\paragraph{Local improvement heuristics}

In this paper, we use a two-level local improvement method with RKGA to improve the optimization results. Level one improvement involves $2$-opt and is applied to all new individuals. Level two involves the Lin-Kernighan (LK) heuristics \cite{Lin1973}, and as the search with LK heuristics is relatively more expensive to perform, the new individuals are thresholded against an individual at $v_{\textrm{thres}}$ percentile of the parent population after level one improvement. If the fitness of the resulting individual is better than the threshold value, the solution is deemed to have more potential and the additional level two improvement step (involving the more expensive the LK heuristic) is then applied to the individuals that passed the threshold.

\section{Experiments and Discussion}

\subsection{Datasets}
\begin{figure}[thpb]
      \centering
      \includegraphics[width=0.44\textwidth]{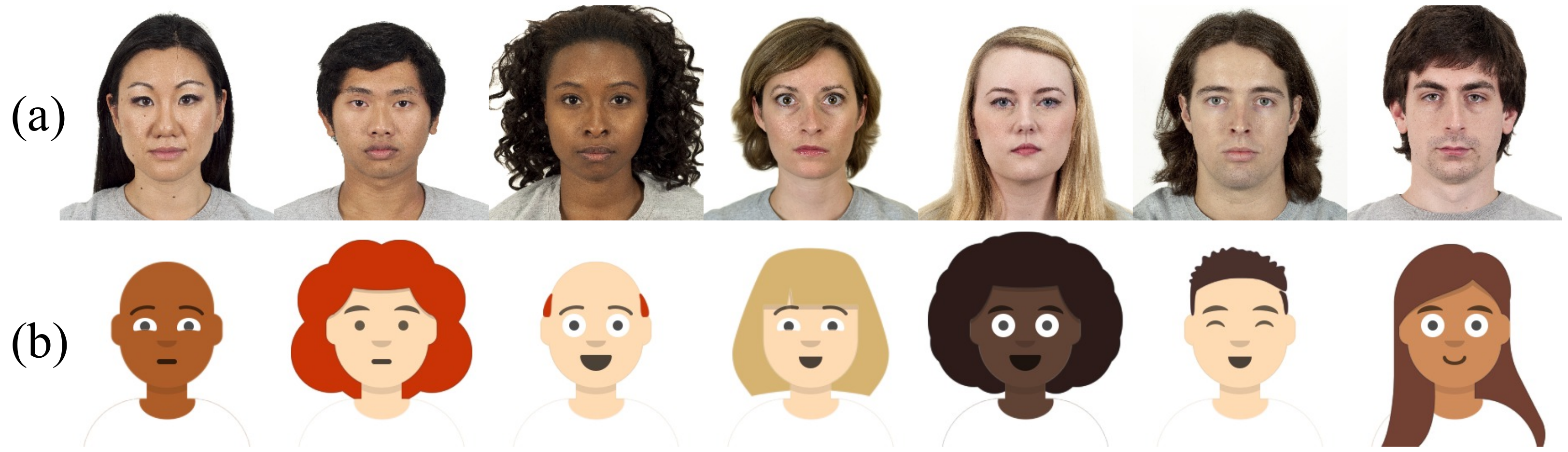}
      \caption{\emph{Unpaired} examples of (a) CFD face images and (b) cartoon-style avatar images.}
      \label{fig:data_example}
\end{figure}

We conducted experiments using the images from Chicago Face Dataset (CFD)~\cite{Ma2015TheCF}. For the cartoon-style avatar image dataset, considering the drawing media of robot arm is marker on the whiteboard, the avatars should be suited to artistic composition with clean, bold lines. In order to meet such a requirement, we used the \textit{Avataaars} library to randomly generate diverse cartoon avatar images as our avatar dataset. Examples of CFD image and generated avatar image are shown in Fig.~\ref{fig:data_example}. We randomly chose 1145 images from the CFD dataset and 852 images from generated avatar dataset to train AvatarGAN.

The training images from CFD dataset and generated by avatar dataset were \emph{unpaired}. These images were firstly converted into grayscale, because the monochromatic sketches were performed by robot. Subsequently, the images were rescaled to $256\times 256$. After that, the facial-area images are directly obtained by simple cropping from fixed position, and then re-scaled to $256\times 256$.

\subsection{Experiments on the Style Transfer Module}

\subsubsection{AvatarGAN Generation}
For AvatarGAN, we used the same architectures of generator and discriminator proposed by CycleGAN for a fair comparison. The generative networks include two convolution networks, nine residual blocks~\cite{He2016DeepRL} and two deconvolution networks with instance normalization~\cite{Ulyanov2016InstanceNT}. All four discriminators utilized $70\times 70$ Patch-GANs~\cite{Li2016PrecomputedRT,Isola2017ImagetoImageTW,ledig2017photo}, which aim to classify whether the $70\times 70$ overlapping image patches are real or fake~\cite{cyclegan2017}. Moreover, we set $\alpha=0.2$ to encourage the generator to focus more on learning facial features. The weight $\lambda$, which controls the relative importance of consistency loss, was set to $10$.

\begin{figure}[ht]
    \centering
    \includegraphics[width=0.45\textwidth]{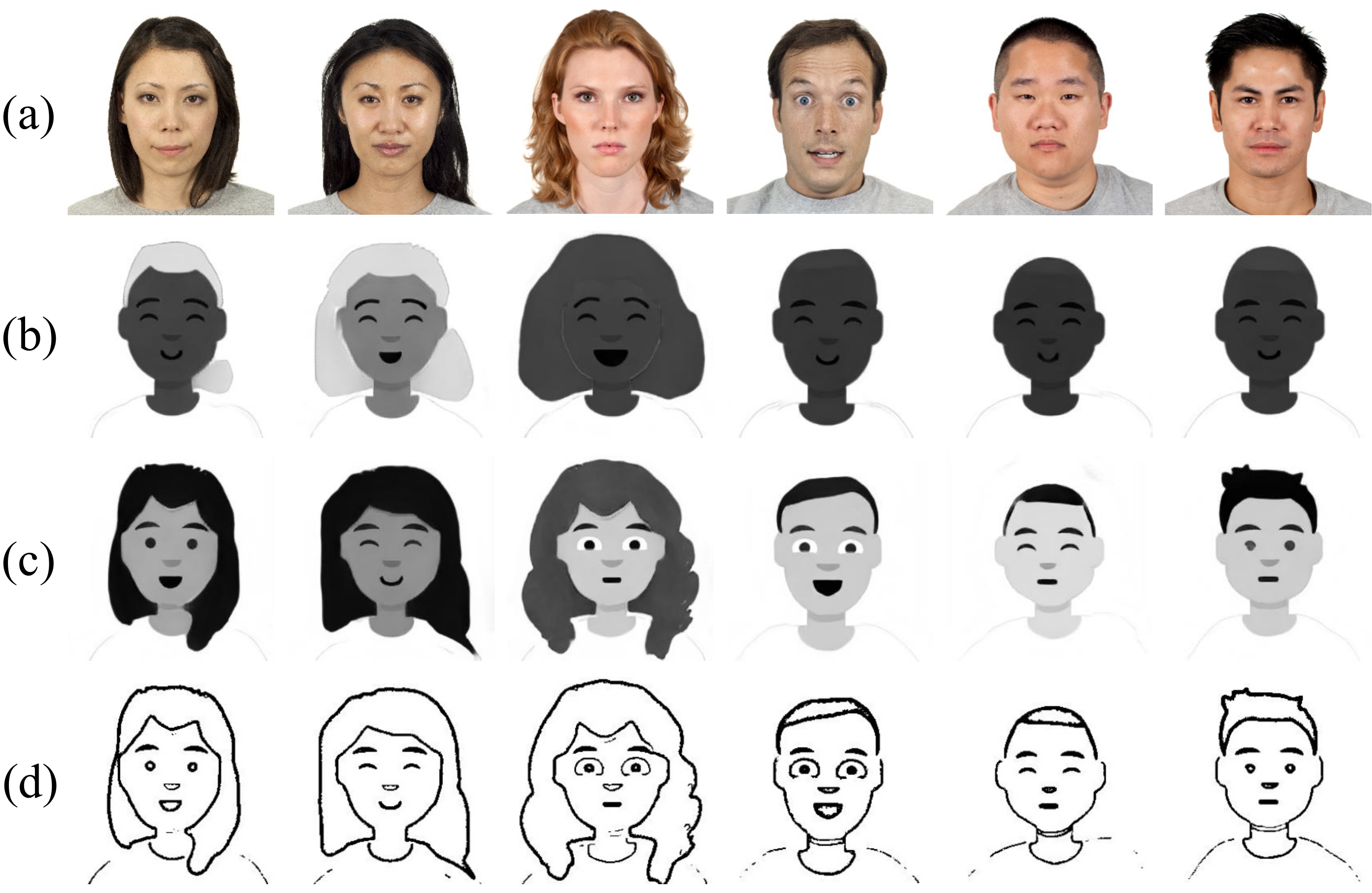}
    \caption{Face-to-avatar translation, where (a) are the real face images, (b) are the transferred results by CycleGAN, (c) are the transferred results by AvatarGAN, and (d) are the coherent contours of (c) generated by FDoG.}
    \label{fig:avatarresult}
\end{figure}

The real face to avatar translation performance of AvatarGAN was evaluated visually first. Fig.~\ref{fig:avatarresult} shows the transferred avatars by CycleGAN (b) and AvatarGAN (c), and the results from coherent contour generation (d). It is observed that the proposed AvatarGAN has two major advantages over CycleGAN in this application. First, AvatarGAN generated avatars with similar facial features as the input images, while CycleGAN seemed to significantly suffer from mode collapse, especially in the facial area. Second, the avatars generated by AvatarGAN are more diverse in the different parts of the cartoon structure. 

It is also shown that CycleGAN fails to preserve important facial features and map them to avatars, since cycle consistency loss only enforces faithful mapping between \emph{whole images}, and does not impose special constraints on any particular areas. In contrast, AvatarGAN preserves and maps facial features from the original faces effectively.

In addition to the above-mentioned advantages,
it is worth noting that AvatarGAN can creatively generate new facial features that are not presented in the original dataset. For instance, there is only one face shape for all the avatars in the dataset (ref. Fig. \ref{fig:data_example}), but the generated avatars have diverse face shapes, such as round, oval, diamond, and square, which are consistent with the original faces. As for another facial feature, the haircut, the resulted haircuts of avatars seem more distinctive and personalized, especially for the long hair. Although the diversity within the training dataset puts constraints on what images could be possibly generated, AvatarGAN still manages to bring in more personalized features and creativity to the avatar generation. 

\subsubsection{Contour Generation}
The contour generation sub-module generates avatar contours as shown in Fig.~\ref{fig:avatarresult} (d). These contours capture important facial features from the generated avatars, such as the haircut, face shape, and facial expressions. In addition, lines in the contours are coherent and smooth, thereby the generated avatar sketches are more suitable to robotic drawing.

\subsubsection{Evaluation of the Generalization}
To evaluate the generalization capability of AvatarGAN, we performed a face-to-avatar translation on additional face images from the CUHK Face Sketch Database (CUFS)

\begin{figure}[htbp]
    \centering
    \includegraphics[width=0.45\textwidth]{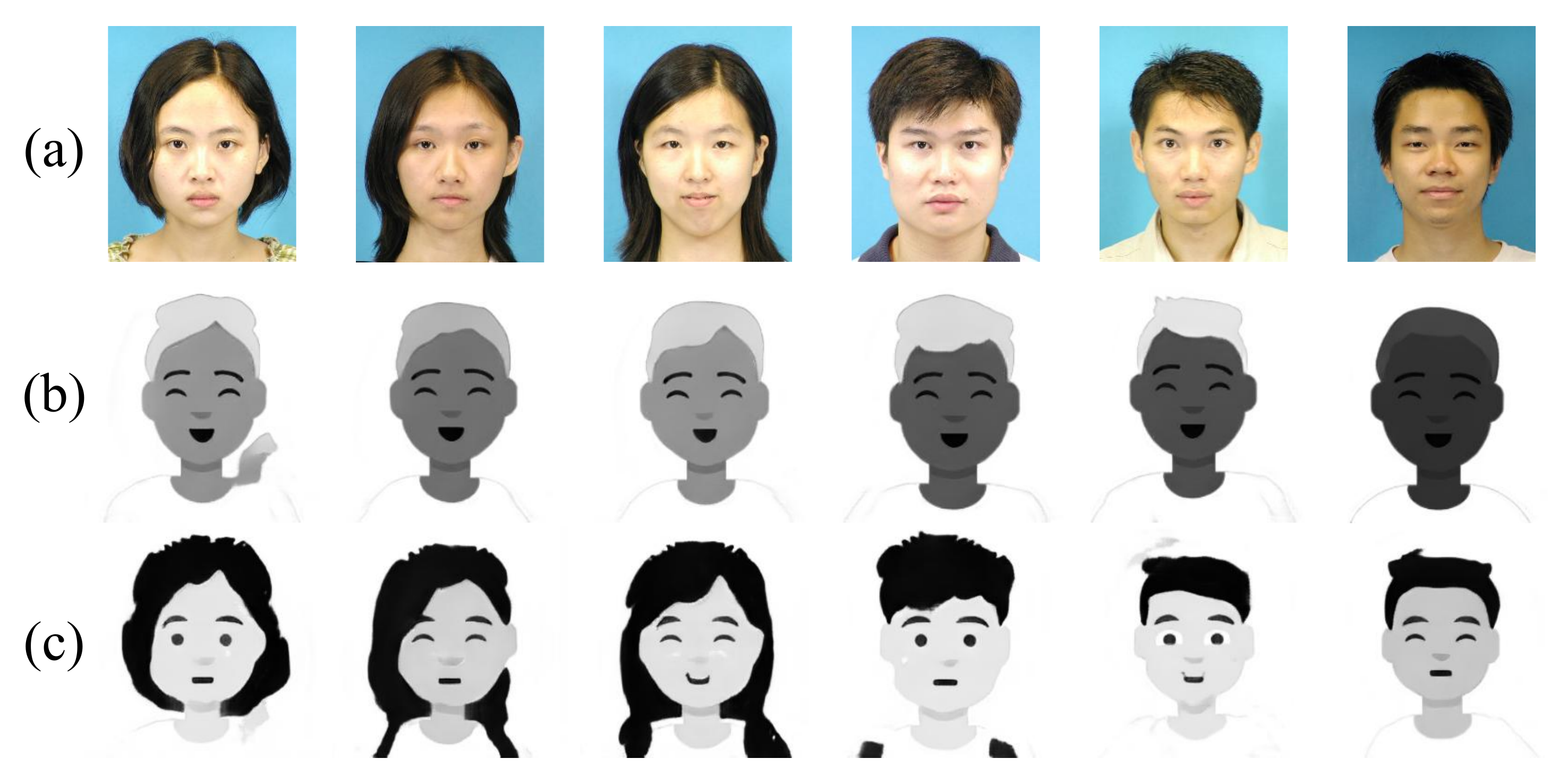}
    \caption{Face-to-avatar translation on the external CUFS dataset, where (a) are the real face images, (b) are the generated cartoon avatars by CycleGAN, and (c) are the translation results by AvatarGAN.}
    \label{fig:avatarresult_new}
\end{figure}

~\cite{wang2009face}. The AvatarGAN and CycleGAN models were trained on the CFD dataset and 
applied to CUFS dataset directly without fine-tune. 
The generated avatars by CycleGAN (b) and AvatarGAN (c) are shown in Fig.~\ref{fig:avatarresult_new}. On the CUFS images, the same observations as on Fig. \ref{fig:avatarresult} were made.

\begin{table}[htbp]
    \caption{Mean cycle consistency loss of CycleGAN and AvatarGAN on the test datasets. The column name indicates the input domain.}
    \label{tab:loss_compare}
    \centering
    \resizebox{\columnwidth}{!}{
    \begin{tabular}{ c c c c c}
        \toprule
        \multirow{2}{*}{Method} & \multicolumn{4}{c}{Mean cycle consistency loss $\mathcal{L}_{cycle}$} \\
        & Face & Avatar & Real facial area & Avatar facial area \\
        \midrule
        CycleGAN & $\mathbf{0.0355}$ & $\mathbf{0.0167}$ & $0.2620$ & $0.3298$ \\
        \midrule
        AvatarGAN & $0.0367$ & $0.0252$ & $\mathbf{0.0329}$ & $\mathbf{0.0252}$ \\
        \bottomrule
    \end{tabular}
    }
\end{table}

Table~\ref{tab:loss_compare} compares the mean cycle consistency loss ($\mathcal{L}_{cycle}$) of CycleGAN and AvatarGAN. AvatarGAN significantly outperformed CycleGAN on the facial-area translation, as the newly introduced facial-area stream (ref. Fig.~\ref{fig:avatar_gan}) enforces the generators of AvatarGAN to preserve facial features in the input image and translate them to the target domain. Surprisingly, CycleGAN slightly outperforms AvatarGAN on the whole-image translation, despite the fact that it generated intermediate images of poor quality and low consistency with input images.
Because CycleGAN sees the whole intermediate image when mapping it back to the input domain, it might exploit some subtle features in non-facial areas to recover the identity of the original image and use the image identity to facilitate the reconstruction of the input image. In other words, CycleGAN may take shortcuts to satisfy cycle consistency, without learning the true mapping of facial features between the two domains. The facial-area translation stream eliminates such shortcuts.

\subsubsection{User Study}
To further validate the effectiveness of the proposed AvatarGAN, we conducted a user study from 10 candidates. In the survey, candidates were asked to match 10 random faces with the paired avatars generated by our system. The mean accuracy from the user study was 98\%, and the standard deviation was 0.04.

\subsection{Experiments on the Robotic Drawing Module}
\subsubsection{Pixel-Coordinates Extraction}
The pixel-coordinate extraction process was tested with a range of avatar image outputs from the style transfer module over 20 test runs. For each test run, the module reliably extracted and obtained the pixel-coordinates of line segments required to construct the robotic line drawing.

\begin{figure}[!htbp]
      \centering
      \includegraphics[width=0.45\textwidth]{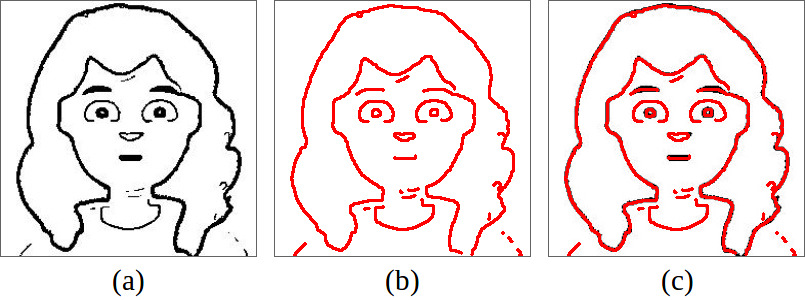}
      \caption{Example of pixel-coordinates extraction. (a) The avatar sketch; (b) Simulated drawing using extracted pixel-coordinates; (c) Overlay of (a) and (b). }
      \label{fig:exp_line_extract}
\end{figure}     

\begin{figure}[!htbp]
    \small
      \centering
      \includegraphics[width=0.45\textwidth]{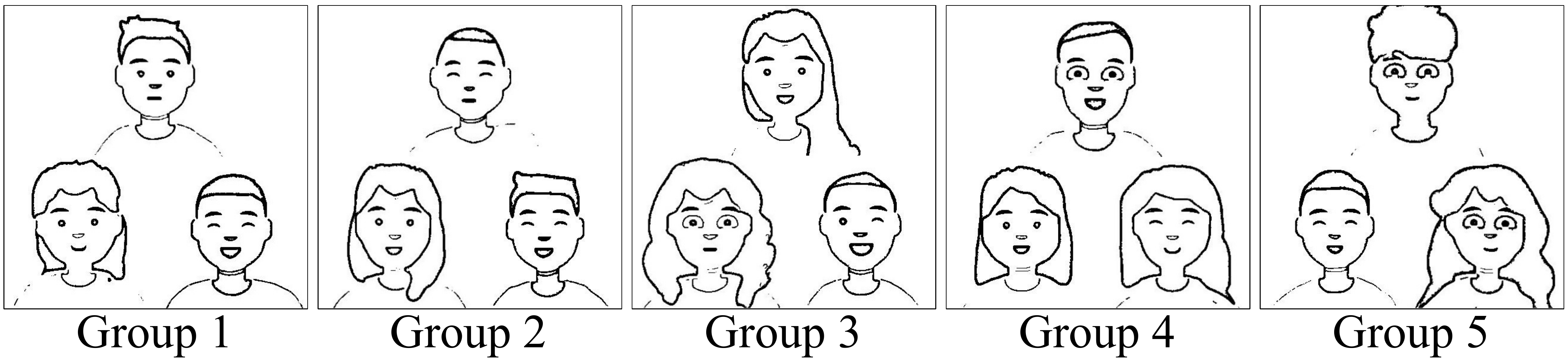}
      \caption{Examples of the avatar groupings used as inputs in the path optimization experiments.}
      \label{fig:exp_pair}
\end{figure}

\subsubsection{Path Optimization}

For additional testing of the path optimization algorithm, we grouped and placed multiple avatars into a single image to increase the complexity of each trial problem. Examples of the images used are shown in Fig.~\ref{fig:exp_pair}. We conducted the experiments across five different avatar groupings (average number of lines to optimize in each image is 74), with 10 trials conducted for each group of avatars.

The parameters used for RKGA are $N = 100$, $r=3$, $p_{\textrm{crossover}}=0.8$, $p_{\textrm{mutation}}=0.5$, and $cost_{\textrm{lift}}=30$. For the local search heuristic, the threshold percentile $v_{\textrm{thres}}$ for the two-level improvement is set by $v_{\textrm{thres}}=\min{(0.05+0.01c,0.10)}$, where $c$ is number of consecutive generations with no improvement in the best solution. For the uniform crossover strategy employed, each key in the individual will originate from the first parent with a probability of $0.7$, with an additional $0.5$ chance for the tour encoded by the first parent to be reversed before the crossover operation. For mutation, the mutated individual will have indexes of the decimal part of the key shuffled with probability $0.05$, followed by a bit-flip for integer part with probability $0.05$.

The proposed optimization algorithm (RKGA w/ 2-opt, LK) was benchmarked against other commonly used methods. We used the greedy search algorithm as a baseline for our comparison, and calculated the percentage improvement of each algorithm's path fitness over the fitness of the greedy path. In particular, we achieved good path optimization results using the RKGA with 2-opt and LK heuristics. The resultant paths had significant improvement against the results from the greedy search method, with an average improvement in path fitness of 17.34\%, as shown in Table \ref{table_gtsp}.

\begin{table}[htbp]
    \caption{Improvements in path fitness value, $v_{\textrm{fitness}}$, of various optimization methods over the greedy benchmark ($\%$).}
    \label{table_gtsp}
    \centering
    \resizebox{\columnwidth}{!}{
    \begin{tabular}{l r r r r r r }
        \toprule
        & G1 & G2 & G3 & G4 & G5 & Avg. \\
        Greedy w/ 2-opt & $16.4$ & $20.2$ & $5.5$ & $8.8$ & $9.8$ & $12.1$ \\
        \midrule
        Greedy w/ 2-opt, LK & $16.9$ & $22.5$ & $14.7$ & $11.0$ & $10.3$ & $15.1$ \\
        \midrule
        RKGA w/ 2-opt & $18.0$ & $22.5$ & $19.8$ & $12.2$ & $11.3$ & $16.8$ \\
        \midrule
        \textbf{RKGA w/ 2-opt, LK} & $\mathbf{19.1}$ & $\mathbf{22.9}$ & $\mathbf{20.4}$ & $\mathbf{12.4}$ & $\mathbf{11.9}$ & $\mathbf{17.3}$ \\
        \bottomrule
    \end{tabular}
    }
\end{table}

\begin{figure}[hbtp]
      \centering
      \includegraphics[width=0.44\textwidth]{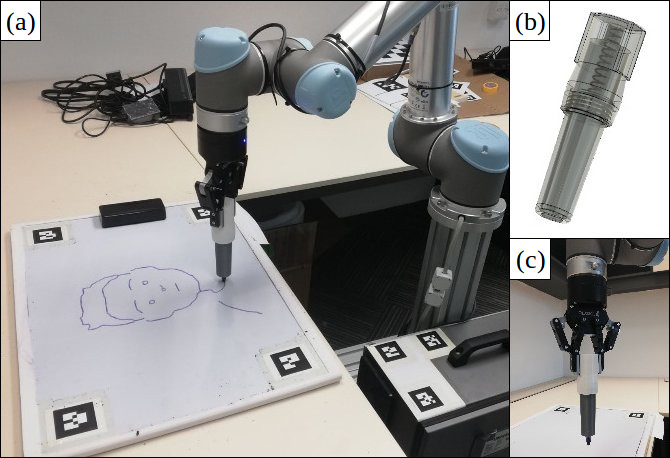}
      \caption{Integrated tests setup. (a) UR5 robot executing a drawing; (b) Tool holder design; (c) Robot end-effector.}
      \label{fig:exp_setup}
\end{figure}

\begin{figure*}[tb]
  \centering
  \includegraphics[width=0.85\textwidth]{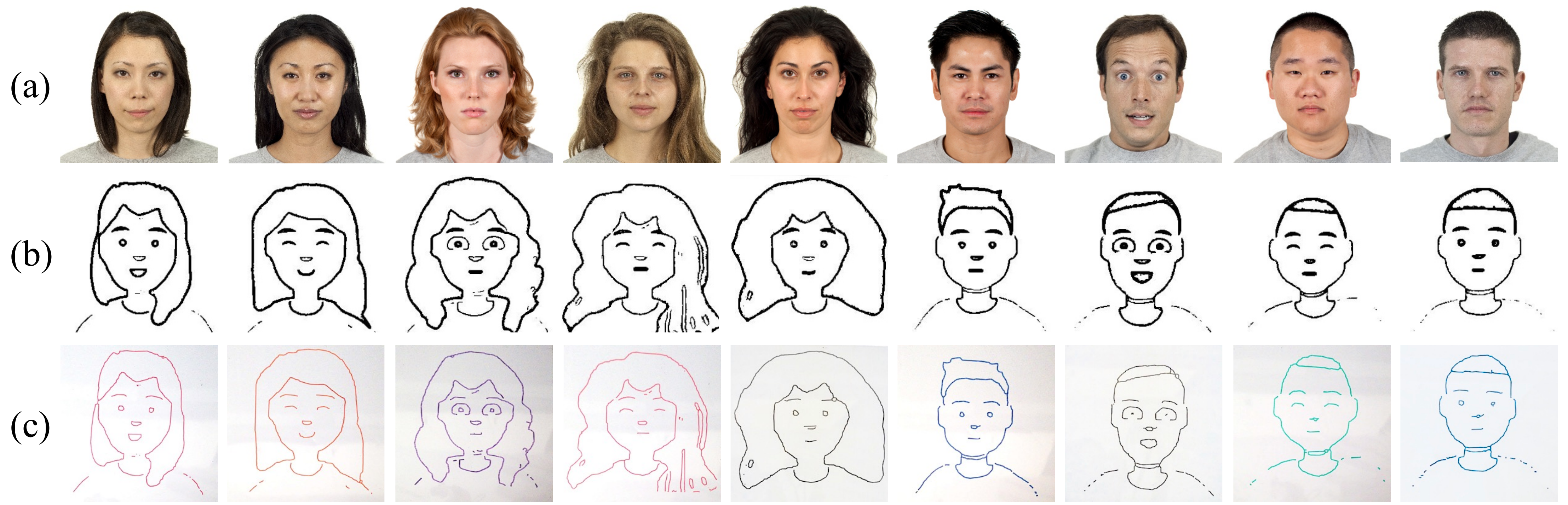}
  \caption{Results of the integrated tests, where (a) are the real face images, (b) are the corresponding cartoon avatars generated by AvatarGAN after coherent contour generation, and (c) are the marker-on-whiteboard drawings finished by the robotic arm.}
  \label{fig:robot_arm_draw}
\end{figure*}

\subsection{Integrated Tests of RoboCoDraw System}

We conducted 20 integrated trials with our RoboCoDraw system. Our drawing system was implemented with the UR5 robotic arm, with a Robotiq two-finger gripper attached as the end-effector. In addition, for more effective gripping and drawing with the marker, a customized spring-loaded marker holder was designed to compensate for small errors in the $z$-axis, ensuring that a constant amount of pressure is applied when drawing on slightly uneven surfaces. Fig.~\ref{fig:exp_setup} shows our experiment setup for robotic drawing. Fig.~\ref{fig:robot_arm_draw} shows examples of the original photographs and the corresponding whiteboard drawings produced in our experiments.

In the integrated tests, the average time used for physical drawing was 43.2 seconds, while the other computational processes (image prepossessing, AvatarGAN translation, contour generation, etc.) took 9.9 seconds to complete.

\section{CONCLUSIONS}

In this paper, we proposed the RoboCoDraw system that facilitates the efficient creation and drawing of personalized avatar sketches on the robotic arm, given real human face images\footnote{Code available at https://github.com/Psyche-mia/Avatar-GAN}. The proposed system consists of 1) a GAN-based style transfer module for personalized, cartoon avatar generation, and 2) an RKGA-based path planning and optimization module, for time-efficient robotic path generation. We compared the two proposed modules with existing methods in the style transfer and path optimization tasks, respectively. For the style transfer, our AvatarGAN generates more diversified cartoon avatars with much better likeness; For the path optimization, our method reduced 17.34\% of the drawing time on average compared with the baseline. The RoboCoDraw system has great potential in public amusement and human-robot interactive entertainment applications.

\section*{Acknowledgements}

\noindent This research is supported by the Agency for Science, Technology and Research (A*STAR) under its AME Programmatic Funding Scheme (Project \#A18A2b0046).

\bibliographystyle{aaai}
\fontsize{9.0pt}{10.0pt} \selectfont
\bibliography{4477_aaai.bib}

\end{document}